\documentclass[10pt, conference, compsocconf, twocolumn]{IEEEtran}
\usepackage[utf8]{inputenc}
\usepackage{amsmath}
\usepackage{graphicx}
\usepackage{booktabs}
\usepackage{hyperref}
\usepackage[superscript]{cite}
\usepackage{tikz}
\usetikzlibrary{arrows.meta, positioning, shapes, fit}
\usepackage{tabularx}
\usepackage{caption}
\usetikzlibrary{arrows.meta, positioning, shapes, fit}
\begin{document}

\title{Structural Gender Bias in Credit Scoring: Proxy Leakage}

\author{
\IEEEauthorblockN{
Navya S. D.$^{1}$ \quad
Sreekanth D.$^{1}$ \quad
Dr. S. S. Uma Sankari$^{2}$
}
\IEEEauthorblockA{
$^{1}$Department of Mathematics, Indian Institute of Space Science and Technology (IIST), Thiruvananthapuram, India\\
$^{2}$Vikram Sarabhai Space Centre (VSSC), Thiruvananthapuram, India
}
}

\maketitle
\thispagestyle{plain}

\begin{abstract}
As financial institutions increasingly adopt machine learning for credit risk assessment, the persistence of algorithmic bias remains a critical barrier to equitable financial inclusion. This study provides a comprehensive audit of structural gender bias within the Taiwan Credit Default dataset, specifically challenging the prevailing doctrine of ``fairness through blindness.'' Despite the removal of explicit protected attributes and the application of industry-standard fairness interventions our results demonstrate that gendered predictive signals remain deeply embedded within non-sensitive features. Utilizing SHAP (SHapley Additive exPlanations), we identify that variables such as Marital Status, Age, and Credit Limit function as potent proxies for gender, allowing models to maintain discriminatory pathways while appearing statistically fair. To mathematically quantify this leakage, we employ an adversarial inverse modeling framework. Our findings reveal that the protected gender attribute can be reconstructed from purely non-sensitive and financial features with an ROC-AUC score of 0.65, proving that traditional fairness audits are insufficient for detecting implicit structural bias. These results advocate for a shift from surface-level statistical parity toward causal-aware modeling and structural accountability in financial AI.
\end{abstract}

\section{Introduction}

The deployment of algorithmic systems in financial services has transitioned from a tool of efficiency to a primary architect of social destiny. While modern credit scoring models are often presented as objective arbiters of creditworthiness, they frequently serve as digital repositories for historical socio-economic disparities \cite{oneil2016}. By codifying past financial behaviors into mathematical certainties, these models risk transforming historical gendered economic disadvantages into permanent barriers to social mobility \cite{benjamin2019}. 

Current regulatory frameworks—including the European Union’s General Data Protection Regulation (GDPR) \cite{eu_gdpr_2016}, the European Union Artificial Intelligence Act (AI Act) \cite{eu_ai_act_2024}, India’s Digital Personal Data Protection Act (DPDPA) of 2023 \cite{india_dpdpa_2023}, and the United States’ Equal Credit Opportunity Act (ECOA) \cite{us_ecoa_1974}—mandate fairness but often rely on the flawed doctrine of ``fairness through blindness.'' Collectively, these statutes seek to protect individual privacy and ensure algorithmic accountability by prohibiting discriminatory treatment based on protected attributes while requiring transparency in automated decision-making processes. This study challenges that doctrine, arguing that the mere exclusion of sensitive attributes is a performative gesture that fails to address the reality of redundant encoding \cite{barocas2016}. We contend that non-sensitive features like credit limits function as sophisticated proxies that allow models to reconstruct protected gender characteristics with high fidelity. 

This research contributes to the discourse on algorithmic accountability by proving that structural bias survives even the most rigorous standard fairness audits. Utilizing the Taiwan Credit Default dataset, we move beyond surface-level statistical parity to expose hidden predictive pathways. Through a combination of SHAP-based explainability and adversarial inverse modeling, we demonstrate that gender information "leaks'' into the model's logic, facilitating a form of implicit discrimination that traditional metrics fail to detect \cite{dwork2012}. Our findings advocate for a shift from statistical fairness toward a structural understanding of AI-driven exclusion.

\section{Related Works}

The transition from traditional scorecards to advanced machine learning frameworks has redefined credit risk assessment, yet it has simultaneously complicated the pursuit of algorithmic equity.

Historically, financial institutions have relied on ``fairness through unawareness''—the intentional exclusion of protected attributes—to satisfy anti-discrimination laws such as the GDPR and ECOA. However, recent researches identifying a ``regulatory paradox'' suggests that this blindness often facilitates systematic mispricing of risk \cite{liu2025}. Specifically, Liu et al. (2025) demonstrate that female borrowers consistently receive lower credit scores than men despite identical risk profiles, suggesting that scoring models are systematically miscalibrated in a manner that penalizes women. This reinforces the foundational argument by Barocas and Selbst (2016) that removing sensitive labels does not eliminate discrimination if the underlying data is a repository of historical socio-economic disadvantage \cite{barocas2016}.

The primary mechanism behind this failure is redundant encoding, where non-sensitive features act as proxies for protected characteristics. Modern research has identified that behavioral patterns of repayment consistency ratios—harbor deeply embedded gendered patterns \cite{cao2024}. Kelly and Mirpourian (2021) highlight that in digital credit underwriting, variables such as mobile hardware specifications and network connection data are so highly correlated with gendered socio-economic status that the omission of an explicit gender attribute is largely performative \cite{kelly2021}. This creates an environment where structural bias survives simple feature exclusion, leading to disparate impacts that traditional auditing tools often overlook.

Current literature categorizes bias mitigation into pre-processing, in-processing, and post-processing interventions \cite{mdpi2025}. While pre-processing techniques like SMOTE and reweighing are widely adopted to address class imbalance, they often fail to address the underlying ``structural bias'' of the feature space. Recent surveys emphasize that while these methods can improve statistical parity, they may do so at the cost of model performance or by introducing new, unquantified biases \cite{garcia2024}. Consequently, there is a maturing research consensus that statistical parity alone is an insufficient metric for ensuring genuine financial inclusion \cite{mdpi2025}.

To bridge the gap between statistical fairness and structural understanding, researchers have turned to Explainable AI (XAI). SHAP (SHapley Additive exPlanations) has emerged as the gold standard for assigning discriminative power to individual features locally and globally \cite{lundberg2017}. Our work extends this forensic approach through adversarial inverse modeling. While traditional XAI explains model outputs, inverse modeling—reconstructing protected attributes from neutral proxies—provides a mathematical measure of the data's inherent bias. This aligns with the ``Fairness Through Awareness'' paradigm, suggesting that true equity requires acknowledging and mitigating the information density of sensitive proxies rather than ignoring them \cite{dwork2012}.

\section{Methodology}
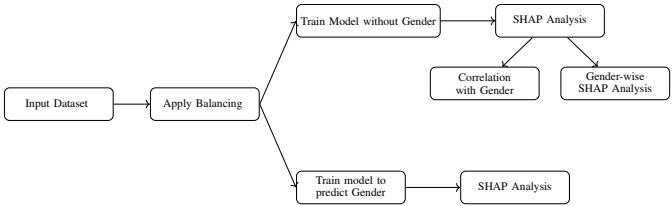
\begin{figure}[t]
\centering
\resizebox{\columnwidth}{!}{%
\begin{tikzpicture}[
    box/.style={
        rectangle,
        rounded corners,
        draw=black,
        thick,
        minimum width=3.0cm,
        minimum height=0.9cm,
        align=center,
        font=\small
    },
    bigbox/.style={
        rectangle,
        rounded corners,
        draw=black,
        thick,
        minimum width=4.8cm,
        minimum height=0.9cm,
        align=center,
        font=\small
    },
    dashedbox/.style={
        rectangle,
        rounded corners,
        draw=black,
        thick,
        dashed,
        inner sep=0.4cm
    },
    arrow/.style={->, thick}
]

\def\Hmain{1.0cm}   
\def\Vbranch{2.3cm} 
\def\Hsmall{1.5cm} 

\node[box] (data) {Input Dataset };
\node[box, right=\Hmain of data] (balance) {Apply Balancing };

\node[box, right=\Hmain of balance, yshift=\Vbranch] (nogender)
{Train Model without Gender};

\node[box, right=\Hmain of balance, yshift=-\Vbranch] (predictgender)
{Train model to\\predict Gender};

\node[box, right=\Hsmall of nogender] (shap1) {SHAP Analysis};
\node[box, right=\Hsmall of predictgender] (shap2) {SHAP Analysis};

\node[box, below=0.8cm of shap1, xshift=-1.8cm] (corr)
{Correlation\\with Gender};

\node[box, below=0.8cm of shap1, xshift=1.8cm] (genderwise)
{Gender-wise\\SHAP Analysis};

\draw[arrow] (data) -- (balance);

\draw[arrow] (balance.east) -- (nogender.west);
\draw[arrow] (balance.east) -- (predictgender.west);

\draw[arrow] (nogender) -- (shap1);
\draw[arrow] (predictgender) -- (shap2);

\draw[arrow] (shap1) -- (corr);
\draw[arrow] (shap1) -- (genderwise);

\end{tikzpicture}%
}
\caption{Methodology Outline}
\label{fig:methodology_outline}
\end{figure}

The methodology of this study is designed to systematically uncover the layers of hidden bias that persist within credit scoring models despite standard fairness interventions. The complete experimental workflow and logical progression of this study are illustrated in Figure \ref{fig:methodology_outline}.

\noindent \textbf{Dataset and Pre-processing : } The study utilizes the Taiwan Credit Default dataset\footnote{The dataset is available at the UCI Machine Learning Repository: \url{https://archive.ics.uci.edu/ml/datasets/default+of+credit+card+clients}}, consisting of 30,000 instances, attributes categorized into non-financial and financial variables. Non-financial features include gender, education, marital status, and age. Financial features encompass the credit limit, monthly repayment status, bill statement amounts, and previous payment amounts. The target variable is binary, representing default payment or non-default. A significant challenge in this dataset is the class imbalance, as only approximately 25\% of records represent defaults. To address this without introducing sampling-induced bias—which could inadvertently accelerate model unfairness by distorting the underlying feature distributions \cite{bao2021}—we implemented three distinct balancing techniques: Class Weighting, Synthetic Minority Over-sampling Technique (SMOTE), and Subsampling. Testing across these diverse methods ensured that the observed bias in our results was structural rather than an artifact of a specific sampling approach.

\noindent \textbf{Experimental Configurations : } We established twelve distinct experimental configurations to evaluate the consistency of results across varying model architectures and data conditions. Two primary machine learning models were employed: Logistic Regression, to provide a linear baseline for sanity-checking explainability results, and XGBoost, to capture complex non-linear structures within the data. Each model was trained using datasets derived from the three sampling techniques previously mentioned—Class Weighting, SMOTE, and Subsampling. Furthermore, each configuration was tested under two feature-set conditions: one including non-financial variables (sensitive attributes) to observe direct influence and potential gender leakage, and another excluding them to investigate whether gendered signals implicitly leak through purely financial variables. This dual-path approach allows for a rigorous validation of gender leakage across both demographic and economic indicators.

\noindent \textbf{Traditional Fairness Audit : } In addition to ensuring optimal predictive accuracy for each machine learning model, we conducted an industry-standard fairness audit to establish a regulatory baseline. The core objective was to demonstrate that a model can satisfy formal fairness criteria while still harboring structural bias. We utilized three primary algorithmic fairness metrics: Disparate Impact, Equalized Odds Difference, and Demographic Parity Difference. Disparate Impact was evaluated against the ideal threshold of 1.0, while both Equalized Odds and Demographic Parity Differences were monitored for proximity to 0.0.

\noindent \textbf{Explainable AI (XAI) and Divergence Analysis : } We integrated SHAP (SHapley Additive exPlanations) to decompose the model's output and assign a contribution value to each feature across all twelve configurations. SHAP was utilized to identify feature importance from a global perspective, allowing us to move beyond black-box accuracy and pinpoint which variables exerted the most significant influence on default predictions. To evaluate how these predictive signals diverge across groups, we conducted separate SHAP analyses for the male and female cohorts, with all values normalized to facilitate a direct and consistent comparison. To mathematically quantify these gender-based logic shifts, we developed a methodology utilizing SHAP-Gender T-statistic analysis. By calculating the statistical variance in feature importance between the two cohorts, we were able to pinpoint exactly where the model's decision-making pathways diverged across protected lines. This granular approach provides empirical evidence of how the model’s internal logic adapts to different demographic signatures, even in the absence of explicit gender attributes.

\noindent \textbf{Inverse Modeling for Bias Validation : } To cross-validate and reinforce our primary findings, we developed an "inverse modeling" framework designed to empirically test for gender leakage. In a truly neutral dataset, any attempt to predict a protected attribute from non-sensitive features should result in near-random performance. We trained secondary "attacker" models to predict the excluded gender attribute using the remaining demographic and financial variables across all twelve experimental configurations. This process was repeated for feature sets both with and without non-financial variables. Including non-financial variables (such as age and marital status) allowed us to quantify the influence of known social stereotypes, while the finance-only configurations tested if gender signals persist within purely economic data. We applied SHAP analysis to these inverse models to identify which specific proxies facilitated the reconstruction of gender information.

\section{Results and Analysis}

\begin{figure}[htbp]
    \centering
    \includegraphics[width=0.5\textwidth]{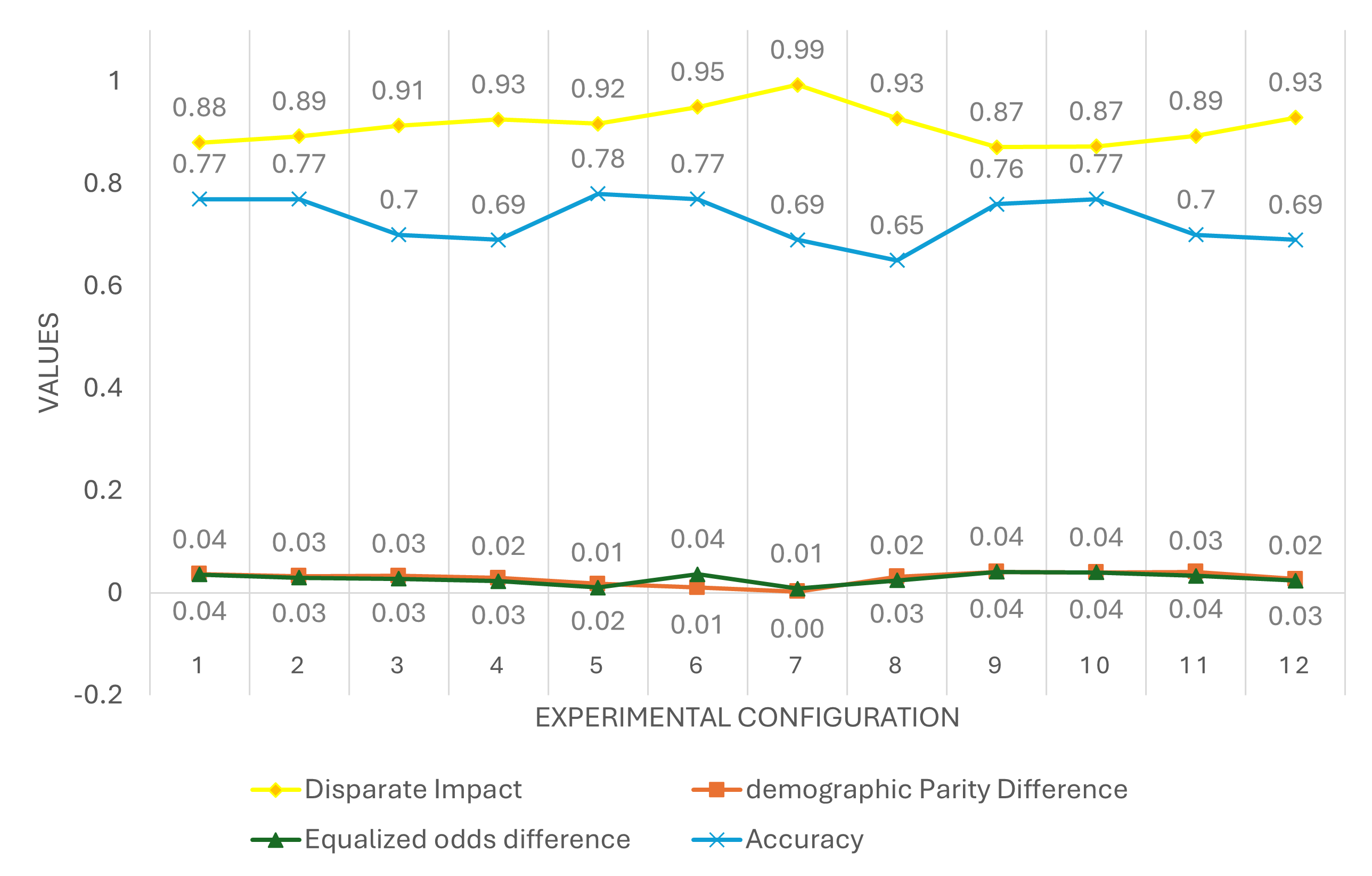}
    \caption{Fairness and accuracy metrics across twelve experimental configurations}
    \label{fig:fairness_metrics}
\end{figure}

The evaluation of the twelve experimental configurations reveals a compelling paradox between surface-level compliance and underlying structural bias. As illustrated in Figure \ref{fig:fairness_metrics}, all models achieve high accuracy levels that reach industry standards, consistently ranging between 0.76 and 0.78. From a traditional regulatory standpoint, these models appear exceptionally fair. We utilized three standard algorithmic tools to measure the fairness of the models: Disparate Impact, Equalized Odds Difference, and Demographic Parity Difference. 

In the ideal case, the Disparate Impact score should be near 1.0, while both the Equalized Odds Difference and Demographic Parity Difference are expected to approach 0.0. As shown in Figure \ref{fig:fairness_metrics}, all twelve configurations satisfy these industry-standard conditions; Disparate Impact scores are near the ideal mark, peaking at 0.99 in Configuration 7, while the Equalized Odds and Parity Difference metrics remain negligible, often near 0.0. These results indicate that the models satisfy all formal fairness conditions. However, the subsequent explainable AI analysis proves that these models still harbor significant hidden bias that traditional tools fail to detect.

\begin{figure}[htbp]
    \centering
    \includegraphics[width=0.5\textwidth]{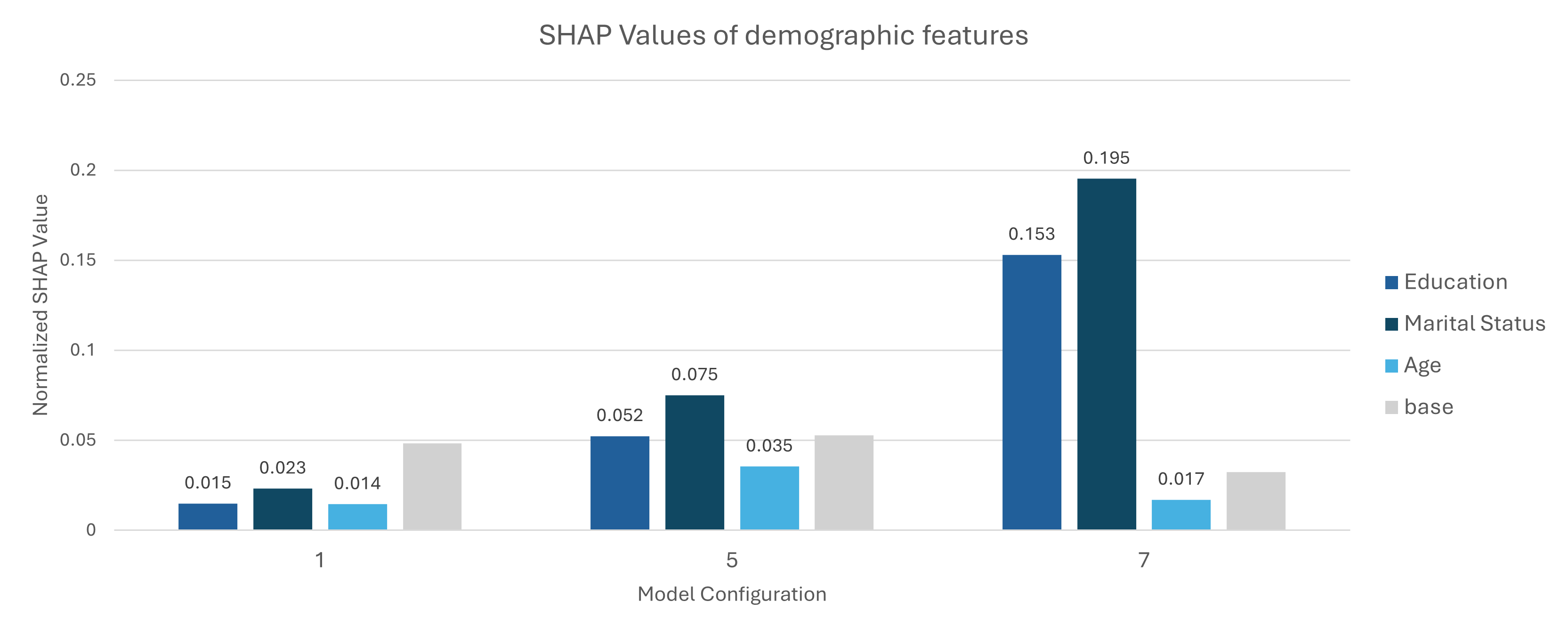}
    \caption{Normalized SHAP values for demographic features in the primary credit default model}
    \label{fig:demographic_shap}
\end{figure}

The primary credit default model exhibits a heavy reliance on demographic features, often surpassing the predictive weight of economic indicators. Figure \ref{fig:demographic_shap} illustrates the normalized SHAP values for models trained on data that explicitly includes sensitive features such as Age, Education, and Marital Status. Notably, all these demographic variables demonstrate feature importance scores higher than the financial variable baseline (indicated as "base" in the figure), where the baseline represents the highest feature importance attributed to any single financial variable in the model. Marital Status consistently emerges as the most influential demographic proxy across configurations, reaching a peak normalized SHAP value of 0.195 in Configuration 7. This comparison provides empirical evidence that whenever sensitive features are included in model training, they exert a greater influence on the final decision-making process than any financial features, thereby fundamentally skewing the model's predictive logic.

To investigate whether gendered signals leak through purely financial indicators, the model was trained exclusively on financial variables, with all sensitive attributes removed. The results, detailed in Table \ref{table:primary_financial_shap}, confirm that PAY\_0 (the history of the most recent repayment status) is consistently the most influential feature in predicting default. A clear temporal trend is observed: as the months decrease from the present, the influence of repayment history progressively weakens. Significant impact is also observed in BILL\_AMT\_1, LIMIT\_BAL, and PAY\_AMT\_1, with their importance slowly decreasing relative to the month of the transaction.

\begin{table}[htbp]
\centering
\resizebox{\columnwidth}{!}{%
\begin{tabular}{@{}lcccc@{}}
\toprule
\textbf{Feature} & \textbf{Config 2} & \textbf{Config 6} & \textbf{Config 8} & \textbf{Config 10} \\ \midrule
LIMIT\_BAL       & 0.11              & 0.07              & 0.02              & 0.10               \\
PAY\_0           & 0.29              & 0.13              & 0.28              & 0.29               \\
PAY\_2           & 0.05              & 0.05              & 0.06              & 0.09               \\
PAY\_3           & 0.05              & 0.05              & 0.03              & 0.05               \\
PAY\_4           & 0.03              & 0.02              & 0.02              & 0.05               \\
BILL\_AMT\_1     & 0.06              & 0.08              & 0.21              & 0.05               \\
BILL\_AMT\_2     & 0.02              & 0.04              & 0.10              & 0.02               \\
PAY\_AMT\_1      & 0.06              & 0.08              & 0.05              & 0.05               \\
PAY\_AMT\_2      & 0.07              & 0.08              & 0.06              & 0.06               \\ \bottomrule
\end{tabular}%
}
\caption{Normalized SHAP Values for Financial Features in Credit Default Prediction}
\label{table:primary_financial_shap}
\end{table}

Notably, LIMIT\_BAL (credit limit) emerges as a dominant factor, often exerting more influence than many of the borrower's current financial activities. Unlike monthly bill payments, which reflect immediate earning capacity, LIMIT\_BAL is a predefined quantity often rooted in historical and structural factors such as income levels, family financial background, and cumulative transaction history. Consequently, the high SHAP values for LIMIT\_BAL prove that the model relies heavily on a feature that depends on a borrower's broader financial and social situation rather than just their absolute present-day payment behavior.

A critical finding of this study is the divergence in how non-sensitive features impact different gender cohorts, indicating that gender information leaks into the model's decision-making process. To understand this further, we performed a T-statistic analysis on the SHAP value differences to measure the variance in feature importance across genders. As detailed in Table \ref{table:demographic_divergence}, Education ($T=2.397$) exerts a significantly higher influence on male default risk than on female risk; specifically, males with a lower education status are pushed toward a default prediction more aggressively than females in the same category.

\begin{table}[htbp]
\centering
\begin{tabularx}{\columnwidth}{@{}l c X@{}}
\toprule
\textbf{Feature} & \textbf{T-Statistic} & \textbf{Observations on Predictive Divergence} \\ \midrule
Education      & 2.397  & Primary driver for males; lower education status increases default risk significantly compared to the female cohort. \\ \addlinespace
Marital Status & -2.093 & Exhibits a stronger predictive signal for female cohort assessment, contributing more heavily to the final default decision. \\ \addlinespace
Age            & -6.976 & Maximum divergence observed; serves as the dominant demographic feature in determining female default status. \\ \bottomrule
\end{tabularx}
\caption{Divergence in Feature Impact by Gender (SHAP-Gender T-Statistics)}
\label{table:demographic_divergence}
\end{table}

Conversely, Marital Status ($T=-2.093$) and Age ($T=-6.976$) demonstrate a more dominant impact on female predictions. In the case of Marital Status, the feature contributes more heavily to the final decision for female borrowers than for their male counterparts. Most notably, Age exhibits the highest divergence across genders, serving as a much stronger predictor for females than for males. This significant variance in Age suggests that the model’s predictive pathways are fundamentally not gender-neutral, as it relies on different demographic "signals" to assess risk for men and women even when explicit gender labels are absent.

\begin{figure}[htbp]
    \centering
    \includegraphics[width=0.5\textwidth]{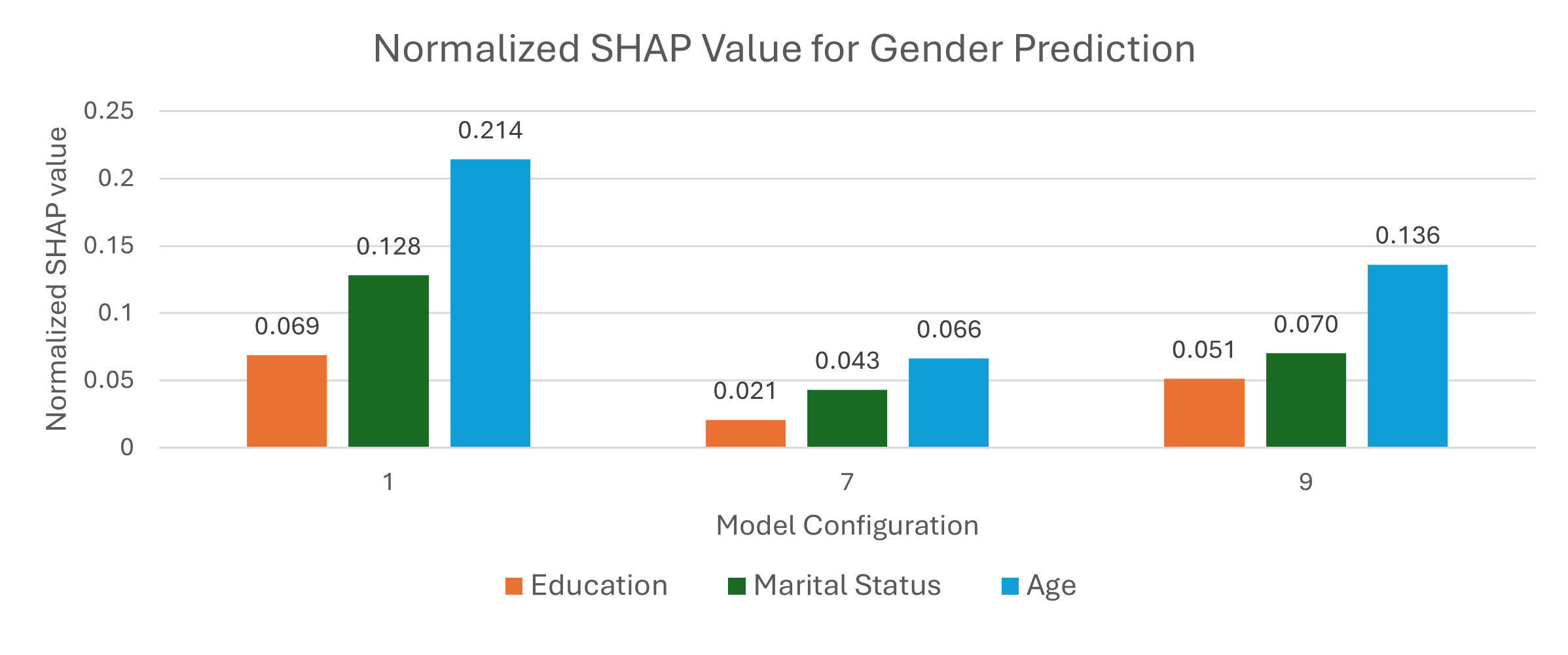}
    \caption{Normalized SHAP values for gender prediction using demographic proxies}
    \label{fig:demographic_leakage}
\end{figure}

The persistence of these implicit signals is further validated by our inverse modeling experiments. The resulting ROC-AUC scores, reaching as high as 0.65, clearly demonstrate that gender prediction is not a random outcome but is driven by a distinct correlation between demographic proxies and gender. As expected, configurations including sensitive attributes such as age and marital status yielded higher predictive accuracy, as these features harbor more potent gender-specific signals and social patterns than purely financial data. As depicted in Figure \ref{fig:demographic_leakage}, Age serves as the primary signal for gender reconstruction (SHAP value up to 0.214), followed by Marital Status. Furthermore, Table \ref{table:financial_leakage} quantifies gender leakage through financial variables. It demonstrates that the Credit Limit (LIMIT\_BAL) and the most recent Bill Amount (BILL\_AMT\_1) carry high gendered signals, with SHAP values reaching 0.199 and 0.220, respectively. 

\begin{table}[htbp]
\centering
\caption{Normalized SHAP Values for Gender Prediction across Financial Proxy Variables}
\label{table:financial_leakage}
\resizebox{\columnwidth}{!}{%
\begin{tabular}{@{}lccccccccc@{}}
\toprule
\textbf{Feature} & \textbf{C1} & \textbf{C2} & \textbf{C3} & \textbf{C4} & \textbf{C6} & \textbf{C9} & \textbf{C10} & \textbf{C11} & \textbf{C12} \\ \midrule
LMT\_BAL     & 0.182 & 0.092 & 0.199 & 0.031 & 0.063 & 0.139 & 0.120 & 0.199 & 0.044 \\
BILL\_AMT\_1 & 0.083 & 0.040 & 0.220 & 0.118 & 0.026 & 0.056 & 0.051 & 0.202 & 0.112 \\
BILL\_AMT\_2 & 0.041 & 0.015 & 0.175 & 0.092 & 0.027 & 0.026 & 0.024 & 0.179 & 0.089 \\
BILL\_AMT\_3 & 0.020 & 0.010 & 0.127 & 0.066 & 0.025 & 0.011 & 0.014 & 0.036 & 0.019 \\
BILL\_AMT\_4 & 0.014 & 0.005 & 0.113 & 0.054 & 0.021 & 0.009 & 0.007 & 0.196 & 0.101 \\
BILL\_AMT\_5 & 0.026 & 0.009 & 0.125 & 0.068 & 0.025 & 0.010 & 0.010 & 0.138 & 0.069 \\ \bottomrule
\end{tabular}%
}
\end{table}

These results provide empirical evidence that sensitive information effectively "leaks" into the model through proxy variables, rendering the "fairness through blindness" strategy insufficient for credit scoring. Features such as Age and Education contain indirect gender data; for instance, distinct distributions within specific age groups for male and female applicants allow the model to predict gender using Age alone, bypassing the need for an explicit gender label. This mechanism reinforces existing social stereotypes embedded in Age and Marital Status. Furthermore, the analysis of financial variables confirms that the two most influential predictors in credit scoring—LIMIT\_BAL and BILL\_AMT\_1—also harbor significant gendered information. As previously established, these are not merely neutral financial metrics; they serve as differentiators that reflect gendered socio-economic realities, thereby facilitating structural gender leakage.

\section{Conclusions and Future Works}

This study provides a comprehensive audit of structural bias in credit scoring models, demonstrating that traditional fairness interventions often fail to address deep-seated gender discrimination. Our results confirm that the "fairness through blindness" approach—removing the protected gender attribute—is a fundamentally flawed strategy based on a blind belief in algorithmic neutrality. We have demonstrated that every sensitive attribute, particularly Age and Marital Status, effectively acts as a proxy for gender. Consequently, even when the explicit gender label is removed, the model learns to discriminate based on these proxies, which in reality reinforces discrimination against the original protected attribute.

Furthermore, we established that removing all sensitive demographic features does not resolve the problem. Financial variables, traditionally viewed as neutral, are not gender-blind; the features with the highest impact on credit scoring, such as Credit Limit and repayment history, carry deeply embedded gender stereotypes. Our inverse modeling experiment provided empirical proof of this "gender leakage," where the model could reconstruct the protected attribute. This proves that trusting algorithms based on surface-level metrics is insufficient.

Moving forward, several critical avenues for research remain to be addressed. First, we aim to develop a formal causal framework to trace and quantify the structural influence of sensitive attributes on model outcomes via indirect pathways. This shift from correlation-based explainability to causal discovery will facilitate a more rigorous understanding of the underlying mechanisms of gender leakage. 

Second, we intend to validate this methodology across a diverse array of international credit datasets. This cross-dataset validation is essential to investigate how localized socio-economic variations and cultural contexts influence the specific proxy variables the model selects for discrimination. By analyzing these localized variations, we can assess the robustness of our detection framework under different regulatory environments. Finally, our future technical work will focus on integrating fairness constraints directly into the model's objective function. We plan to explore causal regularization techniques and the development of fairness-aware loss functions that simultaneously optimize for predictive accuracy, demographic parity, and individual explainability.

\end{document}